\pdfoutput=1

\documentclass[11pt]{article}

\usepackage[]{acl}

\usepackage{times}
\usepackage{url}
\usepackage{caption}
\usepackage{subcaption}
\usepackage{latexsym}
\usepackage{siunitx}
\usepackage{graphicx}
\usepackage{subcaption}
\usepackage{multirow}

\usepackage[T1]{fontenc}

\usepackage[utf8]{inputenc}

\usepackage{microtype}

\title{TIB-VA at SemEval-2022 Task 5:  A Multimodal Architecture for the Detection and Classification of Misogynous Memes}

\author{Sherzod Hakimov$^{1,2}$,  Gullal S. Cheema$^{1}$, \and Ralph Ewerth$^{1,2}$ \\
        $^1$TIB -- Leibniz Information Centre for Science and Technology \\ $^2$Leibniz University Hannover, L3S Research Center \\ Hannover, Germany \\ \{sherzod.hakimov, gullal.cheema, ralph.ewerth\}@tib.eu}

\begin{document}
\maketitle
\begin{abstract}

The detection of offensive, hateful content on social media is a challenging problem that affects many online users on a daily basis. Hateful content is often used to target a group of people based on ethnicity, gender, religion and other factors. The hate or contempt towards women has been increasing on social platforms. Misogynous content detection is especially challenging when textual and visual modalities are combined to form a single context, e.g., an overlay text embedded on top of an image, also known as \textit{meme}. In this paper, we present a multimodal architecture that combines textual and visual features 
in order to detect misogynous meme content. The proposed architecture is evaluated in the \textit{SemEval-2022 Task 5: MAMI - Multimedia Automatic Misogyny Identification} challenge under the team name \textit{TIB-VA}. Our 
solution obtained the best result in the \textit{Task-B} where the challenge is to classify whether a given document is misogynous and further identify the main sub-classes of \textit{shaming}, \textit{stereotype}, \textit{objectification}, and \textit{violence}.

\end{abstract}

\section{Introduction}

Detection of hate speech has become a fundamental problem for many social media platforms such as Twitter, Facebook, and Instagram. There have been many efforts by the research community and companies to identify the applicability of advanced solutions. In general, hate speech is defined as \textit{a hateful language targeted at a group or individuals based on specific characteristics such as religion, ethnicity, origin, sexual orientation, gender, physical appearance, disability or disease}. The hatred or contempt expressed towards women has been drastically increasing, as reported by \citet{misogyny_cases2} and \citet{misogyny_cases}. Detection of such misogynous content requires large-scale automatic solutions \cite{multimodal_classification_visual, multi_off, clicit} and comprehensive annotation processes such as those defined by \citet{misogyny_annotation}.

The detection of hateful content has been mainly studied from the textual perspective based on the Computational Linguistics and Natural Language Processing (NLP) fields. However, hateful content on social media can be found in other forms, such as videos, a combination of text and images, or emoticons. Misogynous content detection is especially challenging when textual and visual modalities are combined in a single context, e.g., an overlay text embedded on top of an image, also known as \textit{meme}. Recent efforts in multimodal representation learning \cite{vilbert, clip} pushed the boundaries of solving such problems by combining visual and textual representations of the given content. Several datasets have been proposed using multimodal data~\cite{mmhs, hateful_memes, memotion, har_meme, multi_off, clicit} for various tasks related to hate speech. Each dataset includes an image and corresponding text, which is either an overlay text embedded on an image or a separate accompanying text such as tweet text. Differently from existing datasets based on memes~\citep{hateful_memes, memotion, har_meme, multi_off} the addressed task aims to identify misogyny in memes specifically. Among the previously mentioned work, only the dataset by \citet{clicit} intended for the detection of misogyny, in which the text is in the Italian language. In terms of the size of the mentioned datasets, the dataset by \citet{mmhs} contains 150K image-text pairs, while other datasets have moderate sizes that range between 2K-10K image-text pairs. Moreover, existing model architectures that are evaluated on such benchmark datasets use a combination of various textual and visual features extracted from pre-trained models \cite{hateful_meme_report}.

\begin{figure*}[h]
	\centering
    \includegraphics[width=0.80\linewidth]{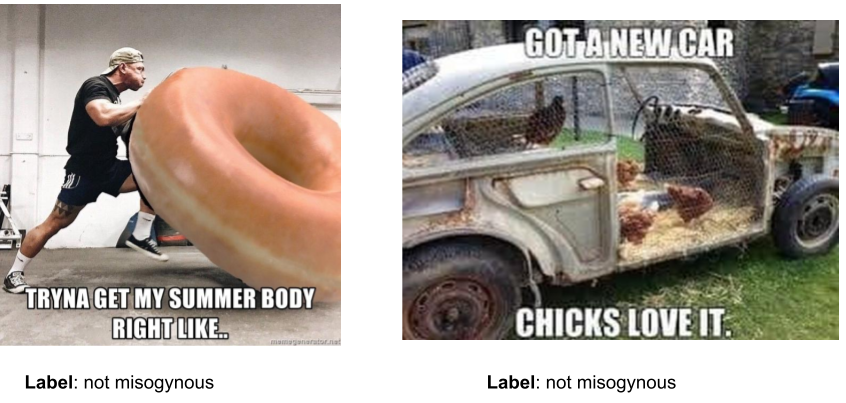}
    \includegraphics[width=0.80\linewidth]{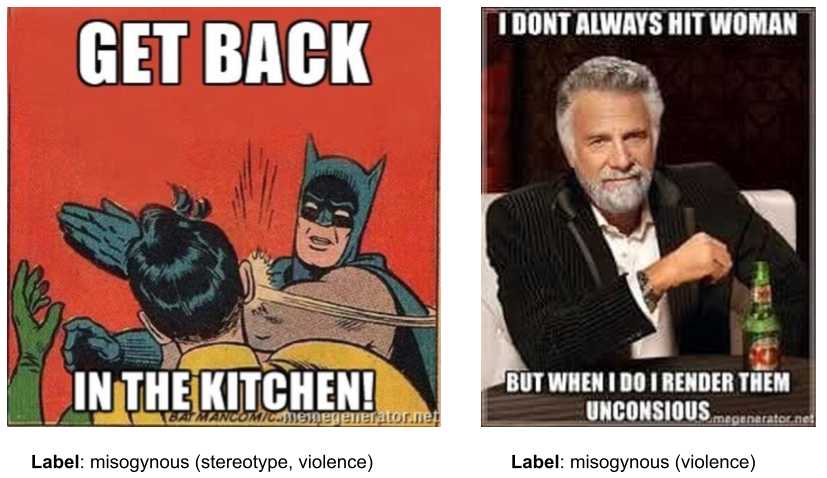}
	\caption{Four data samples from \textit{MAMI - Multimedia Automatic Misogyny Identification} with their corresponding class labels. Misogynous samples have additional sub-classes from \textit{stereotype}, \textit{shaming}, \textit{objectification}, and \textit{violence}.}
	\label{fig:samples}
\end{figure*}

The \textit{SemEval-2022 Task 5: MAMI - Multimedia Automatic Misogyny Identification}~\citep{mami}\footnote{\url{https://competitions.codalab.org/competitions/34175}} is a new challenge dataset that focuses on identifying misogynous memes. The memes in this dataset are composed of an image with an overlay text. Some samples from the dataset with their corresponding class labels are provided in Figure~\ref{fig:samples}. The dataset includes two sub-tasks as described below.

\begin{itemize}
    \item Task-A: a basic task about misogynous meme identification, where a meme should be categorized either as misogynous or not misogynous
    \item Task-B: an advanced task, where the type of misogyny should be recognized among potential overlapping categories such as stereotype, shaming, objectification and violence.
\end{itemize}

In this paper, we present our participating model architecture under the team name \textit{TIB-VA}. The model architecture is based on a neural model that uses pre-trained multimodal features to encode visual and textual content and combines them with an LSTM (Long-short Term Memory) layer. Our proposed solution obtained the best 
result (together with two other teams) on the \textit{Task-B}. 

The remainder of the paper is structured as follows. In Section~\ref{sec:approach}, we describe the proposed model architecture. In Section~\ref{sec:experiments}, the experimental setup, dataset details, as well as evaluations of the model architecture are described in detail. Finally, Section~\ref{sec:conclusion} concludes the paper and outlines areas of future work.

\section{Multimodal Architecture}\label{sec:approach}

\begin{figure*}[h]
	\centering
    \includegraphics[width=1.00\linewidth]{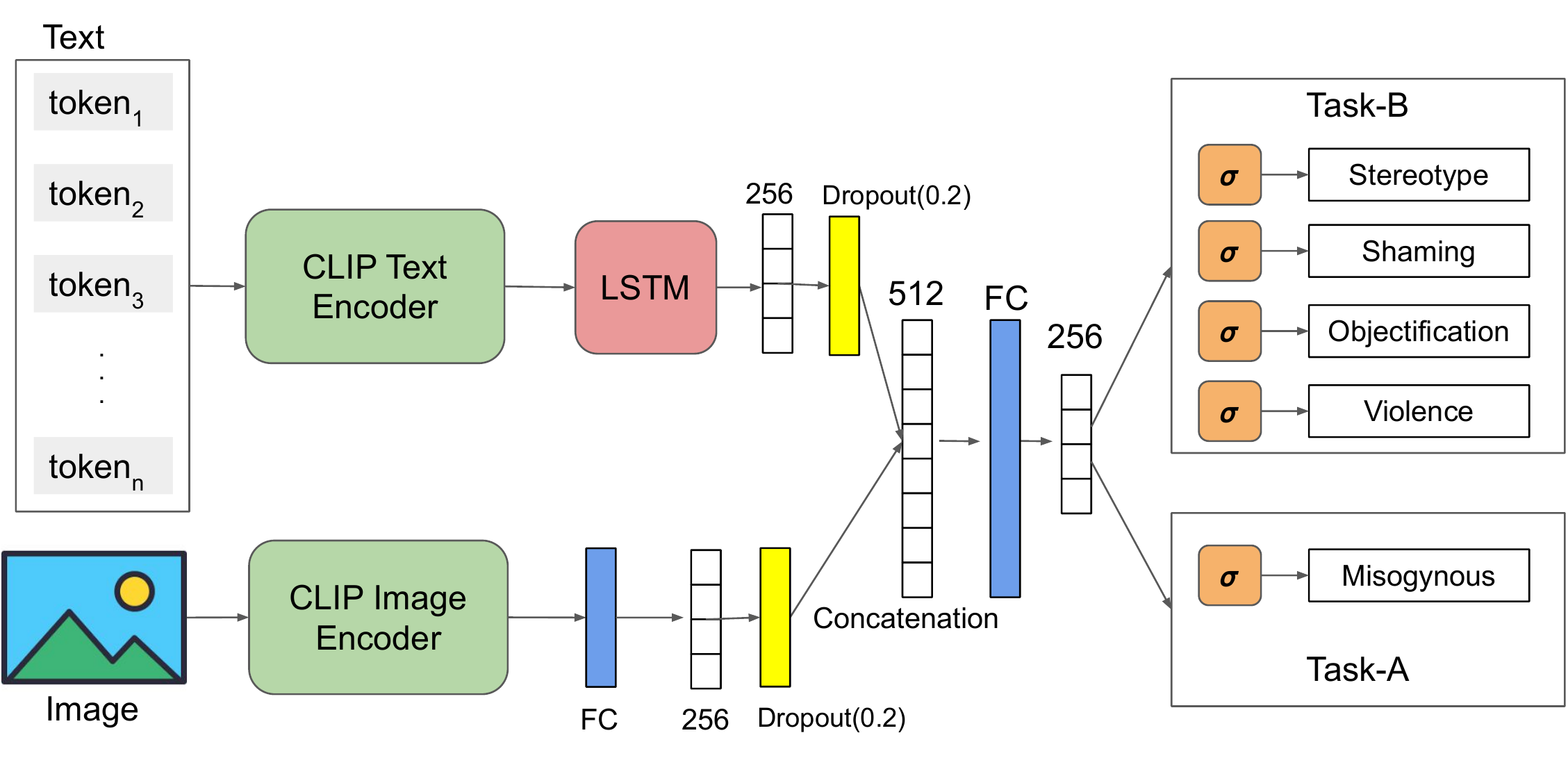}
	\caption{The model architecture that combines textual and visual features to output probabilities for Task-A (misogynous) and Task-B (stereotype, shaming, objectification, violence). FC: Fully connected layer, $\sigma$: sigmoid function.}
	\label{fig:model}
\end{figure*}

Our model architecture is a neural model that uses a pre-trained CLIP~\cite{clip} model to extract textual and visual feature representations. The main reason behind the proposed architecture is the usage of the CLIP model that is pre-trained on over 400 million image-text pairs. We used recently available \textit{ViT-L/14} variant of CLIP. The tokens in the overlay text and the image are fed into \textit{CLIP Text Encoder} and \textit{CLIP Image Encoder} respectively. The text encoder outputs a sequence of \num{768}-dimensional vectors for each input token. These token vectors are then fed into an LSTM layer with a size of \num{256}. This layer is another essential part of the proposed architecture. It learns the contextual relatedness among tokens in the text by combining all token representations extracted from the CLIP text encoder branch. The output from the image encoder is fed into a fully-connected layer with a size of \num{256}. The output from an LSTM layer for text and output from the fully-connected layer for the image are fed into separate dropout layers (dropout rate of \num{0.2}), the outputs are concatenated, and then fed into another fully connected layer with a size of \num{256}. The final vector representation is then fed into separate sigmoid functions for each task. For Task-A, the sigmoid outputs a single value that indicates the probability of misogyny. For Task-B, each sub-class of misogyny (stereotype, shaming, violence, objectification) has a separate sigmoid function that outputs a probability value for the corresponding class. The model architecture is shown in Figure~\ref{fig:model}. The source code of the described model is shared publicly with the community\footnote{\url{https://github.com/TIBHannover/multimodal-misogyny-detection-mami-2022}}.

\section{Experimental Setup and Results}\label{sec:experiments}

\subsection{Dataset}

The \textit{SemEval-2022 Task 5: MAMI - Multimedia Automatic Misogyny Identification}~\citep{mami} aims at identifying misogynous memes by taking into account both textual and visual content. Samples from the dataset are given in Figure~\ref{fig:samples}. The dataset includes the overlay text extracted from an image. The challenge is composed of two sub-tasks. Task-A is about predicting whether a given meme is misogynous or not. Task-B requires models to identify sub-classes of misogyny (stereotype, shaming, violence, objectification) in cases where a given meme is misogynous. The samples in Task-B can have multiple labels where a meme can have a single or all of the above sub-classes of misogyny. The train and test splits have \num{10000} and \num{1000} samples, respectively. The distribution of samples for the corresponding two sub-tasks are given in Table~\ref{tab:dataset_stat}.

\begin{table*}[h]

\centering
\begin{tabular}{|l|cc|cccc|c|}
\hline
\multicolumn{1}{|c|}{\multirow{2}{*}{\textbf{Splits}}} & \multicolumn{2}{c|}{\textbf{Task-A}} & \multicolumn{4}{c|}{\textbf{Task-B}} & \textbf{Total} \\ \cline{2-8} 
\multicolumn{1}{|c|}{} & \multicolumn{1}{l|}{Misogynous} & \multicolumn{1}{l|}{NOT} & \multicolumn{1}{l|}{Shaming} & \multicolumn{1}{l|}{Objectification} & \multicolumn{1}{l|}{Violence} & \multicolumn{1}{l|}{Stereotype} & \multicolumn{1}{l|}{} \\ \hline
\textbf{Train} & \multicolumn{1}{c|}{\num{5000}} & \num{5000} & \multicolumn{1}{c|}{\num{1274}} & \multicolumn{1}{c|}{\num{2202}} & \multicolumn{1}{c|}{\num{953}} & \num{2810} & \num{10000} \\ \hline
\textbf{Test} & \multicolumn{1}{c|}{\num{500}} & \num{500} & \multicolumn{1}{c|}{\num{146}} & \multicolumn{1}{c|}{\num{348}} & \multicolumn{1}{c|}{\num{153}} & \num{350} & \num{1000} \\ \hline
\end{tabular}
\caption{Distribution of samples in Task-A and Task-B for train and test splits in the \textit{MAMI - Multimedia Automatic Misogyny Identification} dataset.}
\label{tab:dataset_stat}
\end{table*}

\subsection{Experimental Setup}

\textbf{Training Process}: The model architecture is trained using Adam optimizer~\cite{DBLP:journals/corr/KingmaB14} with a learning rate of \textit{1e-4}, a batch size of \textit{64} for maximum of \textit{20} epochs. We decrease the learning by half after every five epochs. We use 10\% of the training split for validation to find the optimal hyper-parameters.

\textbf{Implementation}: The model architecture is implemented in Python using the PyTorch library. 

\begin{table}[h]

\centering
\begin{tabular}{|l|c|c|}
\hline
\multicolumn{1}{|c|}{\textbf{Team}} & \textbf{Task-A} & \textbf{Task-B} \\ \hline
Ours (TIB-VA) & \multicolumn{1}{l|}{0.734} & \multicolumn{1}{l|}{\textbf{0.731}} \\ \hline
SRC-B & \multicolumn{1}{l|}{\textbf{0.834}} & \multicolumn{1}{l|}{\textbf{0.731}} \\ \hline
PAFC & \multicolumn{1}{l|}{0.755} & \multicolumn{1}{l|}{\textbf{0.731}} \\ \hline
DD-TIG & \multicolumn{1}{l|}{0.794} & \multicolumn{1}{l|}{0.728} \\ \hline
NLPros & \multicolumn{1}{l|}{0.771} & \multicolumn{1}{l|}{0.720} \\ \hline
R2D2 & \multicolumn{1}{l|}{0.757} & \multicolumn{1}{l|}{0.690} \\ \hline
\end{tabular}
\caption{Experimental results for the selected top-performing teams on the \textit{MAMI} dataset. The results on Task-A and Task-B are Macro-F1 and Weighted F1 measures, respectively.}
\label{tab:results}
\end{table}

\subsection{Results}

The official evaluation results\footnote{\url{https://competitions.codalab.org/competitions/34175\#results}} for the top-performing teams are presented in Table~\ref{tab:results}. The results on Task-A and Task-B are macro-averaged F1 and weighted-averaged F1 measures, respectively. Our model architecture (team \textit{TIB-VA}) achieves the best result (\num{0.731}) on Task-B along with other two teams: \textit{SRC-B} and \textit{PAFC}. The \textit{SRC-B} team has the highest performance on Task-A.

\section{Conclusion}\label{sec:conclusion}

In this paper, we have presented a multimodal model architecture that uses image and text features to detect misogynous memes. The proposed solution is built on the pre-trained CLIP model to extract features for encoding textual and visual content. While the presented solution does not yield top results on Task-A, it achieves the best performance in Task-B for identifying sub-classes of misogyny such as stereotype, shaming, objectification, violence. In future work, we will explore the combination of multiple multimodal features that measure different aspects in visual content such as violence, nudity or specific objects and scene-specific content.

\section*{Acknowledgements}
This work has received funding from the European Union’s Horizon 2020 research and innovation program under the Marie Skłodowska-Curie grant agreement No.~812997 (CLEOPATRA ITN).

\bibliography{references}
\bibliographystyle{acl_natbib}
\end{document}